\documentclass[nohyperref]{article}

\usepackage{microtype}
\usepackage{graphicx}
\usepackage{lipsum}
\usepackage{cuted}
\usepackage{subfigure}
\usepackage{booktabs} % for professional tables
\usepackage[utf8]{inputenc}
\usepackage{newunicodechar}
\usepackage{graphicx}
\newunicodechar{❄}{\includegraphics[scale=0.8]{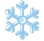}}

\usepackage{hyperref}

\usepackage{multirow}

\usepackage[accepted]{not_icml2022}

\usepackage{amsmath}
\usepackage{amssymb}
\usepackage{mathtools}
\usepackage{amsthm}
\usepackage[font=small]{caption}

\usepackage{pifont}
\newcommand{\cmark}{\ding{51}}%
\newcommand{\xmark}{\ding{55}}%

\usepackage[capitalize,noabbrev]{cleveref}
\crefname{section}{Sec.}{Secs.}
\Crefname{section}{Section}{Sections}
\crefname{figure}{Fig.}{Figs.}
\Crefname{figure}{Figure}{Figures}
\Crefname{table}{Table}{Tables}
\crefname{table}{Tab.}{Tabs.}
\Crefname{appendix}{App.}{App.}

\theoremstyle{plain}

\theoremstyle{definition}

\theoremstyle{remark}

\usepackage[textsize=tiny]{todonotes}

\definecolor{teaser-green}{HTML}{6aa84f}
\definecolor{teaser-blue}{HTML}{4a86e8}
\definecolor{teaser-orange}{HTML}{ff9900}

\usepackage{xspace}
\makeatletter
\DeclareRobustCommand\onedot{\futurelet\@let@token\@onedot}
\def\@onedot{\ifx\@let@token.\else.\null\fi\xspace}
\def\eg{e.g\onedot} 
\def\ie{i.e\onedot} 
 
\def\etc{etc\onedot}

\makeatother

\icmltitlerunning{PaLM-E: An Embodied Multimodal Language Model}

\begin{document}

\twocolumn[{%

\icmltitle{\vspace{-1.1cm}PaLM-E: An Embodied Multimodal Language Model}

\icmlsetsymbol{equal}{*}
\vspace{-0.1cm}
\begin{icmlauthorlist}
\icmlauthor{Danny Driess}{robotics,berlin}
\icmlauthor{Fei Xia}{robotics}
\icmlauthor{Mehdi S. M. Sajjadi}{google}
\icmlauthor{Corey Lynch}{robotics}
\icmlauthor{Aakanksha Chowdhery}{google} \\
\icmlauthor{Brian Ichter}{robotics}
\icmlauthor{Ayzaan Wahid}{robotics}
\icmlauthor{Jonathan Tompson}{robotics}
\icmlauthor{Quan Vuong}{robotics}
\icmlauthor{Tianhe Yu}{robotics}
\icmlauthor{Wenlong Huang}{robotics}
\icmlauthor{Yevgen Chebotar}{robotics}
\icmlauthor{Pierre Sermanet}{robotics}
\icmlauthor{Daniel Duckworth}{google}
\icmlauthor{Sergey Levine}{robotics}
\icmlauthor{Vincent Vanhoucke}{robotics}
\icmlauthor{Karol Hausman}{robotics}
\icmlauthor{Marc Toussaint}{berlin}
\icmlauthor{Klaus Greff}{google}
\icmlauthor{Andy Zeng}{robotics}
\icmlauthor{Igor Mordatch}{google}
\icmlauthor{Pete Florence}{robotics}\\[0.2cm]
$^1$Robotics at Google \ \ $^2$TU Berlin \ \ \ $^3$Google Research \\[0.1cm]
\url{https://palm-e.github.io}
\end{icmlauthorlist}

\icmlkeywords{Robotics, Vision-Language Models, Language Models, Machine Learning}

\vskip 0.2in

\begin{center}
    \includegraphics[width=0.99\linewidth,trim={0 0 0 3mm},clip]{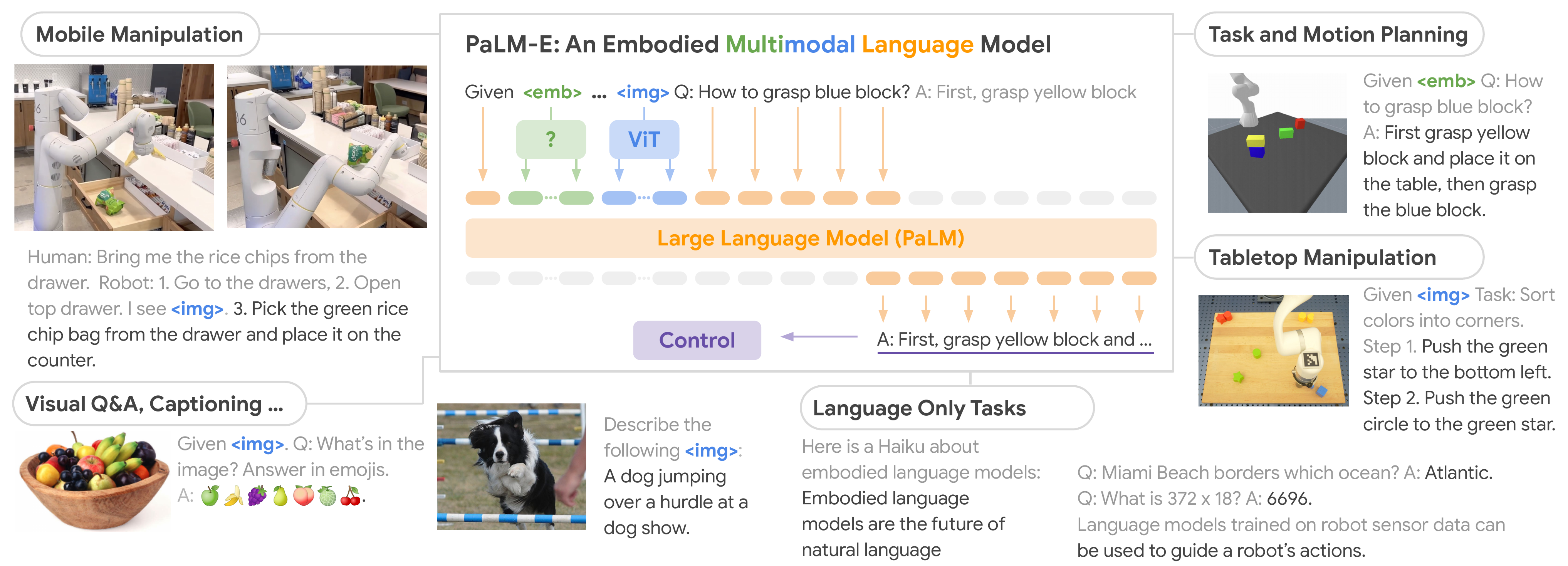}
    \vspace{-0.5em}
    \captionof{figure}{\small{PaLM-E is a single general-purpose multimodal language model for embodied reasoning tasks, visual-language tasks, and language tasks. PaLM-E \textit{transfers} knowledge from visual-language domains into embodied reasoning -- from robot planning in environments with complex dynamics and physical constraints, to answering questions about the observable world. PaLM-E operates on \textit{multimodal sentences}, \ie sequences of tokens where inputs from arbitrary modalities (\eg images, neural 3D representations, or states, in {\color{teaser-green}green} and {\color{teaser-blue}blue}) are inserted alongside text tokens (in {\color{teaser-orange}orange}) as input to an LLM, trained end-to-end.}}
    \label{fig:approach-diagram}
    \vspace{0.5em}
\end{center}

}]

\begin{abstract}
\em{
Large language models have been demonstrated to perform complex tasks. However, enabling general inference in the real world, e.g.\ for robotics problems, raises the challenge of grounding. We propose embodied language models to directly incorporate real-world continuous sensor modalities into language models and thereby establish the link between words and percepts. Input to our embodied language model are multi-modal sentences that interleave visual, continuous state estimation, and textual input encodings. We train these encodings end-to-end, in conjunction with a pre-trained large language model, for multiple embodied tasks including sequential robotic manipulation planning, visual question answering, and captioning.
Our evaluations show that PaLM-E, a single large embodied multimodal model, can address a variety of embodied reasoning tasks, from a variety of observation modalities, on multiple embodiments, and further, exhibits positive transfer: the model benefits from diverse joint training across internet-scale language, vision, and visual-language domains.
Our largest model, PaLM-E-562B with 562B parameters, in addition to being trained on robotics tasks, is a visual-language generalist with state-of-the-art performance on OK-VQA, and retains generalist language capabilities with increasing scale.
%
%In addition to our focus of using PaLM-E as an embodied reasoner, we also describe various innovations that may be of interest in general multimodal learning: the use of neural scene representations as particularly effective input modalities, and text-labeling multimodal tokens for flexible multimodal-grounded textual reasoning.  These options are evaluated, in concert, with large-scale co-training on visual-language datasets.
%- across multiple embodiments
}
\end{abstract}

\section{Introduction}
\vspace{-2mm}

\begin{figure*}[h]
  \centering
  \includegraphics[width=1.0\linewidth]{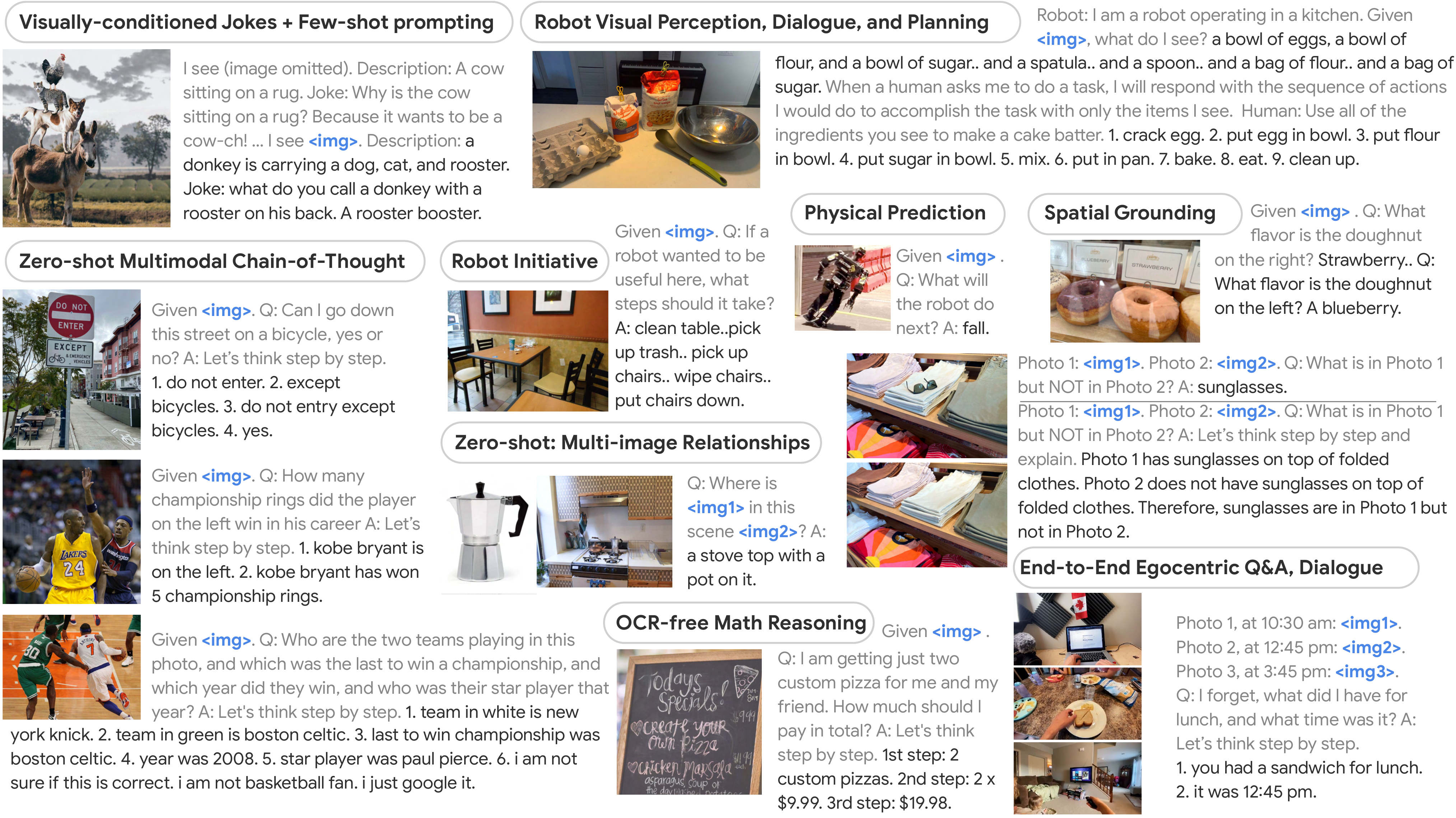}  
 \vspace{-1.5em}
  \caption{\small{
PaLM-E-562B can do {\em{zero-shot multimodal chain-of-thought reasoning}}, can tell visually-conditioned jokes given an image, and demonstrates an array of robot-relevant multimodal-informed capabilities including perception, visually-grounded dialogue, and planning.  PaLM-E also generalizes, zero-shot, to multi-image prompts despite only being trained on single-image prompts.  PaLM-E can also perform math given an image with textually-interleaved handwritten numbers. In addition, the model can perform, zero-shot, question and answering on temporally-annotated egocentric vision, similar to what was shown in \cite{zeng2022socratic} but end-to-end all in one model.
  }}
  \vspace{-1.0em}
  \label{fig:figure2}
\end{figure*}

Large language models (LLMs) demonstrate strong reasoning capabilities across various domains, including dialogue \cite{glaese2022improving,thoppilan2022lamda}, step-by-step reasoning \cite{wei2022chain,kojima2022large}, math problem solving \cite{lewkowycz2022solving,polu2022formal}, and code writing \cite{chen2021evaluating}.
However, a limitation of such models for inference in the real world is the issue of
grounding: while training LLMs on massive textual
data may lead to representations that relate to our
physical world, \emph{connecting} those representations \emph{to} real-world visual
and physical sensor modalities is essential to solving a wider range of \emph{grounded} real-world
problems in computer vision and robotics \cite{tellex2020robots}.
Previous work \citep{ahn2022can} interfaces the output of LLMs with learned robotic
policies and affordance functions to make decisions,
but is limited in that the LLM itself is only provided with textual
input, which is insufficient for many tasks where the geometric configuration of the scene is important.
Further, in our experiments we show that current state-of-the-art \emph{visual}-language models trained on typical vision-language tasks such as visual-question-answering (VQA) cannot directly solve robotic reasoning tasks.

In this paper we propose embodied language models, which
directly incorporate continuous inputs from sensor modalities of an
embodied agent and thereby enable the language model \emph{itself} to
make more grounded inferences for sequential decision making in the real world.
Inputs such as images and state estimates are embedded into the same latent embedding
as language tokens and processed by the self-attention layers of a
Transformer-based LLM in the same way as text.
We start from a pre-trained LLM in which we inject the continuous inputs through an encoder.
These encoders are trained end-to-end to output sequential decisions in terms of natural text that can be interpreted by the embodied agent by conditioning low-level policies or give an answer to an embodied question.
We evaluate the approach in a variety of settings, comparing different input representations (e.g.\ standard vs.\ object-centric ViT encodings for visual input), freezing vs.\ finetuning the language model while training the encoders, and investigating whether co-training on multiple tasks enables transfer.

To investigate the approach's breadth, we evaluate on three robotic manipulation domains (two of which are closed-loop in the real-world), standard visual-language tasks such as VQA and image captioning, as well as language tasks.
Our results indicate that multi-task training improves performance compared to training models on individual tasks.
We show that this \emph{transfer} across tasks can lead to high data-efficiency for robotics tasks, e.g.\ significantly increasing learning success from handfuls of training examples, and even demonstrating one-shot or zero-shot generalization to novel combinations of objects or unseen objects.

We scale PaLM-E up to 562B parameters, integrating the 540B PaLM \cite{chowdhery2022palm} LLM and the 22B Vision Transformer (ViT) \cite{dehghani2023scaling} into, to our knowledge, the largest vision-language model currently reported.
PaLM-E-562B achieves state-of-the-art performance on the OK-VQA \cite{okvqa} benchmark, without relying on task-specific finetuning. Although not the focus of our experimentation, we also find (Fig.~\ref{fig:figure2}) that PaLM-E-562B exhibits a wide array of capabilities including zero-shot multimodal chain-of-thought (CoT) reasoning, few-shot prompting, OCR-free math reasoning, and multi-image reasoning, despite being trained on only single-image examples. Zero-shot CoT \cite{kojima2022large}, originally a language-only concept, has been shown on {\em{multimodal}} data with task-specific programs \cite{zeng2022socratic} but to our knowledge, not via an end-to-end model.

To summarize our main contributions, we (1) propose and demonstrate that a generalist, transfer-learned, multi-embodiment decision-making agent can be trained via mixing in embodied data into the training of a multimodal large language model. We show that, (2) while current state-of-the-art general-purpose visual-language models out-of-the-box (zero-shot) do not well address embodied reasoning problems, it is possible to train a competent general-purpose visual-language model that is also an efficient embodied reasoner.  In studying how to best train such models, we (3) introduce novel architectural ideas such as neural scene representations and entity-labeling multimodal tokens.  Finally, in addition to our focus on PaLM-E as an embodied reasoner we (4) show that PaLM-E is also a quantitatively competent vision and language generalist, and (5) demonstrate that scaling the language model size enables multimodal finetuning with less catastrophic forgetting.

\vspace{-1mm}

\section{Related Work}\label{sec:relatedWork}

\textbf{General vision-language modeling.}
Building on successes in large language \cite{brown2020language, devlin2018bert} and vision \cite{dosovitskiy2020image} models, recent years have seen a growing interest in large vision-language models (VLMs) \cite{li2019visualbert,lu2019vilbert,hao2022language,gan2022vision}.
Unlike their predecessors, VLMs are capable of simultaneously understanding both images and text, and can be applied to tasks such as visual question answering \cite{zhou2020unified,zellers2021merlot}, captioning \cite{hu2022scaling}, optical character recognition \cite{li2021trocr}, and object detection \cite{chen2021pix2seq}. 
The methods by which images are integrated varies.
For example, \citet{alayrac2022flamingo} augments pretrained language models with a mechanism to directly attend to a single context image. 
In contrast, PaLM-E represents images and text as ``multimodal sentences'' of latent vectors, allowing it to process multiple images in a flexible way within any part of a sentence.
More closely related to our work is Frozen~\citep{tsimpoukelli2021multimodal} where vision encoder parameters are optimized via backpropagation through a frozen LLM~\citep{lu2021pretrained}. 
Inspired by this work, we investigate the design in a broader scope by introducing alternative input modalities (\eg neural scene representations), and our proposed approach empirically outperforms Frozen by more than $45\%$ on the VQAv2 benchmark.
More importantly we demonstrate that PaLM-E is applicable not only to perceptual but also embodied tasks.

\textbf{Actions-output models.}
Prior works focus on combining vision and language inputs in an embodied setting with the goal of direct action prediction~\cite{guhur2022instruction,shridhar2022perceiver,shridhar2022cliport,zhang2021hierarchical,silva2021lancon,jang2022bc,nair2022learning,lynch2022interactive,brohan2022rt}.
Among these methods, VIMA~\citep{jiang2022vima} explores multimodal prompts similar to PaLM-E.
The role of language is perhaps most aptly described as task specification in these works.
In contrast, PaLM-E generates high-level instructions as text; in doing so, the model is able to naturally condition upon its own predictions and directly leverage the world knowledge embedded in its parameters.
This enables not only embodied reasoning but also question answering, as demonstrated in our experiments.
Among works that output actions, perhaps most similar is the approach proposed in Gato~\cite{reed2022generalist} which, like PaLM-E, is a generalist multi-embodiment agent. 
In contrast to Gato, we demonstrate positive transfer across different tasks where the model benefits from diverse joint training across multiple domains.

\textbf{LLMs in embodied task planning.}
There have been several methods proposed to leverage LLMs in embodied domains.
While many works focus on understanding natural language \emph{goals} \cite{lynch2020language, shridhar2022cliport, nair2022learning, lynch2022interactive}, fewer consider natural language as a representation for \emph{planning} -- the focus of this work.
LLMs contain vast amounts of internalized knowledge about the world \cite{bommasani2021opportunities}, but without grounding, generated plans may be impossible to execute.
One line of research has employed prompting to elicit a sequence of instructions directly from an LLM either by leveraging semantic similarity between an LLM's generation and an eligible set of instructions~\cite{huang2022language}, incorporating affordance functions~\cite{ahn2022can}, visual feedback~\cite{huang2022inner}, generating world models~\cite{nottingham2023embodied,zellers2021piglet}, planning over graphs and maps~\cite{shah2022lm,huang2022visual}, visual explanations~\cite{wang2023describe}, program generation~\cite{liang2022code,singh2022progprompt}, or injecting information into the prompt~\cite{zeng2022socratic}.
In contrast, PaLM-E is trained to generate plans directly without relying on auxiliary models for grounding.
This in turn enables direct integration of the rich semantic knowledge stored in pretrained LLMs into the planning process.

\vspace{-1mm}
With few exceptions, the parameters of the LLMs employed in many of these works are employed as-is without further training.
In LID~\cite{li2022pre}, this constraint is relaxed and LLM parameters are finetuned to produce a planning network for generating high-level instructions.
$\text{(SL)}^3$ \cite{sharma2021skill} tackles the more challenging task of simultaneously finetuning two LLMs: a planning network, which produces high-level instructions, and a low-level policy network, which selects actions.
With PaLM-E, our interests are distinct and complementary: we investigate a generalist, multi-embodiment model, across multiple modalities.

\section{PaLM-E: An Embodied Multimodal Language Model}\label{sec:embodyingLLMs}
The main architectural idea of PaLM-E is to inject continuous, embodied observations such as images, state estimates, or other sensor modalities into the language embedding space of a pre-trained language model.
This is realized by encoding the continuous observations into a sequence of vectors with the same dimension as the embedding space of the language tokens.
The continuous information is hence injected into the language model in an analogous way to language tokens. PaLM-E is a decoder-only LLM that generates textual completions autoregressively given a prefix or prompt. We call our model PaLM-\textbf{E}, since we use PaLM \cite{chowdhery2022palm} as the pre-trained language model, and make it \textbf{E}mbodied.

\vspace{-1mm}
The \emph{inputs} to PaLM-E consist of text and (multiple) continuous observations. The multimodal tokens corresponding to these observations are interleaved with the text to form \emph{multi-modal sentences}.
An example of such a multi-modal sentence is \texttt{\small Q: What happened between <img\_1> and <img\_2>?} where \texttt{\small <img\_$i$>} represents an embedding of an image.
The \emph{output} of PaLM-E is text generated auto-regressively by the model, which could be an answer to a question, or a sequence of decisions produced by PaLM-E in textual form that should be executed by a robot. When PaLM-E is tasked with producing decisions or plans, we assume that there exists a low-level policy or planner that can translate these decisions into low-level actions. Prior work has discussed a variety of ways to train such low-level policies~\citep{lynch2020language,brohan2022rt}, and we use these prior methods directly without modification.
In the following, we describe our approach more formally.

\textbf{Decoder-only LLMs.}
Decoder-only large language models (LLMs) are generative models trained to predict the probability $p(w_{1:L})$ of a piece of text $w_{1:L} = (w_1, \ldots, w_L)$ that is represented as a sequence of tokens $w_i\in\mathcal{W}$.
Typical neural architectures realize this by factorizing into
% \vspace{-1mm}
\begin{align}
	p(w_{1:L}) = \prod_{l=1}^{L}p_\text{LM}(w_l|w_{1:l-1}), \label{eq:LLM}
% \vspace{-1mm}
\end{align}
where $p_\text{LM}$ is a large transformer network.

\textbf{Prefix-decoder-only LLMs.}
Since the LLM is auto-regressive, a pre-trained model can be conditioned on a prefix $w_{1:n}$ without the necessity to change the architecture
% \vspace{-1mm}
\begin{align}
	p(w_{n+1:L}|w_{1:n}) = \prod_{l=n+1}^{L}p_\text{LM}(w_l|w_{1:l-1}).
% \vspace{-1mm}
\end{align}
The prefix or \emph{prompt} $w_{1:n}$ provides the context based on which the LLM continues to predict the subsequent tokens $w_{n+1:L}$.
This is often used for inference to steer the predictions of the model.
For example, the prompt can contain a description of the task the LLM should solve or examples of desired text completions for similar tasks.

\textbf{Token embedding space.}
The tokens $w_i$ are elements of a fixed vocabulary $\mathcal{W}$ which is a discrete, finite set corresponding to (sub)words in natural language.
Internally, the LLM embeds $w_i$ into a word token embedding space $\mathcal{X}\subset\mathbb{R}^k$ via $\gamma : \mathcal{W}\rightarrow \mathcal{X}$, \ie $p_\text{LM}(w_l|x_{1:l-1})$
%\begin{align}
%	p(w_{n+1:L}|x_{1:n})
%	= \prod_{l=n+1}^{L}p_\text{LM}(w_l|x_{1:l-1})
%\end{align}
with $x_i = \gamma(w_i)\in\mathbb{R}^k$.
The mapping $\gamma$ is typically represented as a large embedding matrix of size $k\times|\mathcal{W}|$ and 
trained end-to-end.
In our case, $|\mathcal{W}|=256\,000$ \cite{chowdhery2022palm}.

\textbf{Multi-modal sentences: injection of continuous observations.}
Multi-modal information such as image observations can be injected into the LLM by skipping the discrete token level and directly mapping the continuous observations into the language embedding space $\mathcal{X}$.
To this end, we train an encoder $\phi : \mathcal{O}\rightarrow \mathcal{X}^q$ that maps a (continuous) observation space $\mathcal{O}$ (refer to \cref{sec:inputRepresentations} for details) into a \emph{sequence} of $q$-many vectors in $\mathcal{X}$.
These vectors are then interleaved with normal embedded text tokens to form the prefix for the LLM. This means that each vector $x_i$ in the prefix is formed from either the word token embedder $\gamma$ or an encoder $\phi_i$:
\vspace{-1mm}
\begin{align}
	\!x_i = \begin{cases}
		\gamma(w_i) & \text{if $i$ a is text token, or}\\
		\phi_j(O_j)_{i} & \text{if $i$ corresponds to observation $O_j$}.\!
	\end{cases}
\vspace{-1mm}
\end{align}
Note that a single observation $O_j$ is usually encoded into multiple embedding vectors.
It is possible to interleave different encoders $\phi_i$ at different locations in the prefix to combine, e.g., information from different observation spaces.
Injecting the continuous information this way into the LLM reuses its existing positional encodings.
In contrast to other VLM approaches (e.g, \cite{chen2022pali}), the observation embeddings are not inserted at fixed positions, but instead placed dynamically within the surrounding text.

\textbf{Embodying the output: PaLM-E in a robot control loop.}
PaLM-E is a generative model producing text based on multi-model sentences as input.
In order to connect the output of the model to an embodiment, we distinguish two cases.
If the task can be accomplished by outputting text only as, \eg, in embodied question answering or scene description tasks, then the output of the model is directly considered to be the solution for the task.

Alternatively, if PaLM-E is used to solve an embodied planning or control task, it generates text that conditions low-level commands.
In particular, we assume to have access to policies that can perform low-level skills from some (small) vocabulary, and a successful plan from PaLM-E must consist of a sequence of such skills.
Note that PaLM-E must determine on its own which skills are available based on the training data and the prompt, and no other mechanism is used to constrain or filter its outputs.
Although these policies are language conditioned, they are not capable of solving long-horizon tasks or taking in complex instructions.
PaLM-E is hence integrated into a control-loop, where its predicted decisions are executed through the low-level policies by a robot, leading to new observations based on which PaLM-E is able to replan if necessary.
In this sense, PaLM-E can be understood as a high-level policy that sequences and controls the low-level policies.

\vspace{-3mm}
\section{Input \& Scene Representations for Different Sensor Modalities}
\vspace{-1mm}
\label{sec:inputRepresentations}
In this section, we describe the individual modalities that we incorporate into PaLM-E, and how we set up their encoders. We propose different architectural choices for each encoder $\phi : \mathcal{O}\rightarrow\mathcal{X}$ to map the corresponding modality into the language embedding space.
We investigate state estimation vectors, Vision Transformers (ViTs)~\cite{dosovitskiy2020image,chen2022pali,ryoo2021tokenlearner} for 2D image features, and the 3D-aware Object Scene Representation Transformer (OSRT)~\cite{sajjadi2022osrt}.
In addition to encoders that represent the input scene globally, we consider object-centric representations that factor observations into tokens that represent individual objects in the scene.

\textbf{State estimation vectors.}
State vectors, \eg from a robot or a state estimate for objects, are perhaps the simplest to input into PaLM-E. Let $s\in\mathbb{R}^S$ be a vector describing the state of the objects in a scene.
For example, $s$ could contain the pose, size, color \etc of those objects.
Then, the MLP $\phi_\text{state}$ maps $s$ into the language embedding space.

\textbf{Vision Transformer (ViT).}
ViT $\tilde{\phi}_\text{ViT}$ \cite{dosovitskiy2020image} is a transformer architecture mapping an image $I$ into a number of token embeddings $\tilde{x}_{1:m} = \tilde{\phi}_\text{ViT}(I)\in\mathbb{R}^{m\times\tilde{k}}$.
We consider several variants, including the 4 billion parameter model from \citet{chen2022pali}, which we refer to as ViT-4B, and a similar 22 billion parameter model, ViT-22B \cite{dehghani2023scaling}, both of which have been pre-trained on image classification.
We further investigate the ViT token learner architecture (ViT + TL) \cite{ryoo2021tokenlearner} which is trained end-to-end from scratch.
Note that the dimensionality $\tilde{k}$ of the ViT embeddings is not necessarily the same as that of the language model.
We therefore project each embedding into $x_{i} = \phi_\text{ViT}(I)_i = \psi(\tilde{\phi}_\text{ViT}(I)_i)$ with $\psi$ being a learned affine transformation.

\textbf{Object-centric representations.}
Unlike language, visual input is not pre-structured into meaningful entities and relationships: while ViT may capture semantics, the structure of the representation resembles a static grid rather than a collection of object instances.
This poses a challenge both for interfacing with LLMs which have been pre-trained on symbols, and for solving embodied reasoning which requires interaction with physical objects.
We therefore also explore structured encoders that aim to separate visual inputs into distinct objects before injecting them into the LLM.
Given ground-truth object instance masks $M_j$, we can decompose ViT's representation into $x_{1:m}^j = \phi_\text{ViT}(M_j\circ I)$ for object $j$.

\textbf{Object Scene Representation Transformer (OSRT).}
An alternative that does not require ground-truth segmentations is OSRT~\cite{sajjadi2022osrt}: rather than relying on external knowledge about objects, they are discovered in an unsupervised way through inductive biases in the architecture \cite{locatello2020object}.
Based on SRT~\cite{sajjadi2022scene}, OSRT learns 3D-centric neural scene representations on in-domain data through a novel view synthesis task.
Its scene representations consist of object slots $o_j = \bar{\phi}_\text{OSRT}(I_{1:v})_j\in\mathbb{R}^{\bar{k}}$.
We project each of these slots into $x_{1:m}^j = \psi(\bar{\phi}_\text{OSRT}(I_{1:v})_j)$ with an MLP $\psi$.
Note that individual objects are always tokenized into \emph{multiple} embeddings each, \ie $\psi : \mathbb{R}^{\bar{k}}\rightarrow\mathbb{R}^{m\times k}$ for OSRT maps into $m$-many embeddings.

\textbf{Entity referrals.}
For embodied planning tasks, PaLM-E must be able to reference objects in its generated plan.
In many cases, including the majority of our experiments, objects in a scene can be identified in natural language by some of their unique properties.
However, there also exist settings where objects are not easily identifiable by language in few words, \eg if there are multiple blocks on a table of the same color at different locations.
For object-centric representations such as OSRT, we label the multi-modal tokens corresponding to an object in the input prompt as follows: \texttt{\small Object 1 is <obj\_1>.
% Object 2 is <obj\_2>.
$\ldots$~Object $j$ is <obj\_$j$>.}
This enables PaLM-E to reference objects via special tokens of the form \texttt{\small obj\_$j$} in its generated output sentences.
In this case, we assume that the low-level policies operate on these tokens as well.

\vspace{-2mm}
\section{Training Recipes}
\vspace{-1mm}
PaLM-E is trained on a dataset of the form $D = \left\{\left(I_{1:u_i}^i, w_{1:L_i}^i, n_i \right)\right\}_{i=1}^N$, where each example $i$ consists of $u_i$-many continuous observations $I_j^i$, a text $w_{1:L_i}^i$, and an index $n_i$.
Despite being a decoder-only model, the text consists of a prefix part up to index $n_i$ that is formed from multi-modal sentences, and the prediction target, which only contains text tokens.
The loss function is therefore a cross-entropy loss averaged over the individual non-prefix tokens $w_{n_i+1:L_i}^i$.
To form the multi-modal sentences within the model, we have special tokens in the text that get replaced by the embedding vectors of the encoders at the locations in the text of those tokens.
We base PaLM-E on the pre-trained 8B, 62B, and 540B parameter variants of PaLM as the decoder-only LLM into which we inject the continuous observations through the input encoders.
Those encoders are either pre-trained or trained from scratch, see Sec.~\ref{sec:inputRepresentations}.
We refer to an 8B LLM combined with a 4B ViT as PaLM-E-12B, similarly a 62B LLM + 22B ViT as PaLM-E-84B, and 540B LLM + 22B ViT as PaLM-E-562B.

\textbf{Variation with Model freezing.}
Most of our architectures consist of three parts, an encoder $\tilde{\phi}$, a projector $\psi$, and the LLM $p_\text{LM}$.
When training PaLM-E, one way is to update the parameters of all these components.
However, LLMs show impressive reasoning capabilities if supplied with a suitable prompt \cite{wei2022chain}.
Therefore, we investigate whether it is possible to \emph{freeze} the LLM and to just train the input encoders, and if so, how different-modality encoders compare.
In this case, the encoder has to produce embedding vectors such that the frozen LLM is grounded on the observations, and also propagate information to the LLM about the capabilities of an embodiment.
Training such encodings can be understood as a form of input-conditioned soft-prompting \cite{tsimpoukelli2021multimodal}, in relation to normal soft prompts \cite{lester2021power}.
In experiments with $\phi_\text{OSRT}$, we also freeze the slot representation, \ie we only update the small projector $\psi$ which serves as the interface between OSRT and the LLM.

\textbf{Co-training across tasks.}
In our experiments, we investigate the effects of co-training our models on a variety of diverse data.  The ``full mixture'',
% cf.~Sec.~\ref{sec:app:dataMixture},
see \cref{sec:app:dataMixture}, consists primarily of a diverse set of internet-scale vision-and-language data, from a variety of tasks.
The sampling frequencies are set such that only 8.9\% of the full mixture is embodied data, and there are several tasks for each embodiment.

\vspace{-2mm}
\section{Experiments}
\vspace{-1mm}
Our experiments consider diverse robotic (mobile) manipulation tasks across three different robot embodiments, in simulation and with two different real robots.
We refer to \url{https://palm-e.github.io} for videos showing the capabilities of PaLM-E on those tasks.
Although not the focus of our work, we evaluate PaLM-E also on general vision-language tasks such as visual-question-answering (VQA), image captioning, and established language modeling tasks.

\begin{figure}[h]
  \centering
  \hspace{-1.5em}
  \includegraphics[width=1.04\linewidth]{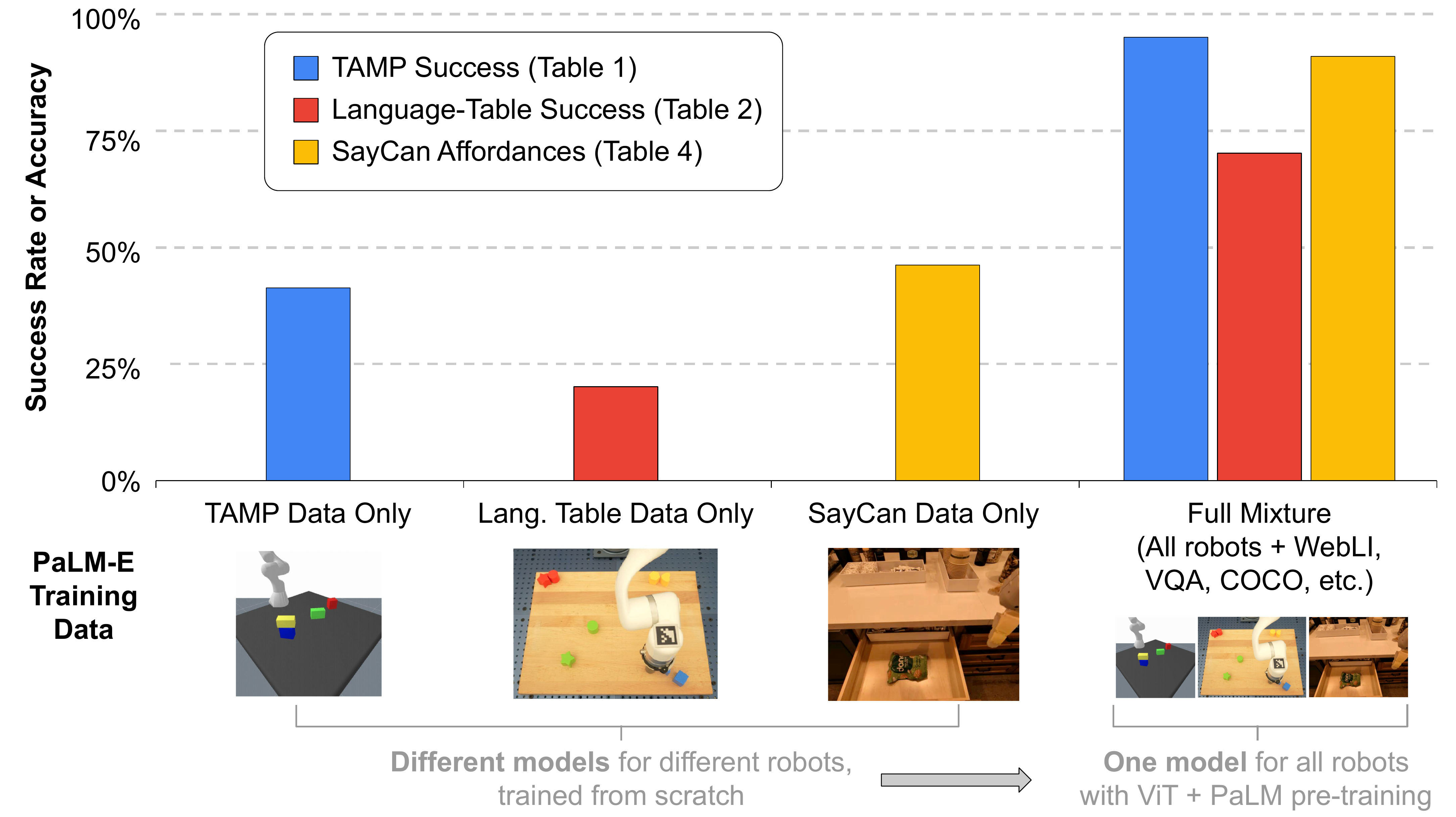}  
 %\vspace{0.0em}
  \caption{\small{
Overview of {\em{transfer}} learning demonstrated by PaLM-E: across three different robotics domains, using PaLM and ViT pretraining together with the full mixture of robotics and general visual-language data provides a significant performance increase compared to only training on the respective in-domain data. See Tab.~\ref{tab:damp_one_percent_data}, Fig.~\ref{fig:damp_transfer}, Tab.~\ref{tab:lt_sim}, Tab.~\ref{table:fractal_sd} for additional data in each domain.
  }}
  \vspace{-1.0em}
  \label{fig:transfer}
\end{figure}

We split our experimental investigation into two broad categories.
First, we compare the different input representations from Sec.~\ref{sec:inputRepresentations} with respect to performance, generalization, and data-efficiency.
The second thread of experiments focuses on one architecture, the main PaLM-E version, consisting of a pre-trained ViT and PaLM language model that takes in raw images as the continuous inputs.
Here we show that a single model, trained on a mixture of many datasets, across diverse tasks, and across robot embodiments, can simultaneously achieve high performance on all of those tasks.
Crucially, we investigate whether co-training on these datasets enables \emph{transfer} (Fig.~\ref{fig:transfer}): despite different tasks and embodiments, the performance on the individual tasks increases by training on the mixture of tasks.
We study the influence on performance, generalization, and data efficiency with respect to co-training strategies and  model parameter size.
Finally, we consider if freezing the LLM and just training the ViT that injects vision into the LLM is a viable path.

As baselines, we consider the state-of-the art visual language model PaLI \cite{chen2022pali}, which has not been trained on embodiment robot data, as well as the SayCan algorithm \cite{ahn2022can}, supplied with oracle affordances.

\vspace{-1mm}
\subsection{Robot Environments / Tasks}
\vspace{-1mm}
Our three robot environments (Fig.~\ref{fig:approach-diagram}) include
a Task and Motion Planning (TAMP) domain where a robot has to manipulate (grasp and stack) objects, a table-top pushing environment, and a mobile manipulation domain.
In each domain, PaLM-E is trained on expert data from that domain.
In many cases, this is a sparse amount of data per task.
The TAMP tasks involve large combinatorics over possible plans, and many decision sequences are infeasible.
PaLM-E has to generate plans that consist of multiple steps, with complicated decision boundaries.
The multi-object tabletop pushing environment is taken from the publicly available Language-Table dataset \cite{lynch2022interactive} and is challenging since it includes several objects, large cardinality of language, and complex pushing dynamics.
For both the TAMP and Language-Table environment, PaLM-E has to reason about the poses of the objects.
It is not sufficient to know which objects are on the table or knowing their rough relationships, the more fine-grained details about the scene geometry are important for solving the tasks.
Finally, we consider a mobile manipulation domain similar to SayCan \cite{ahn2022can}, where a robot has to solve a variety of tasks in a kitchen environment, including finding objects in drawers, picking them, and bringing them to a human.
For all domains we consider both planning and VQA tasks in those environments.
For the mobile manipulation and Language-Table environments, PaLM-E is integrated into the control loop to execute the plans in the real world, and has to adjust the plan in presence of external disturbances or failures of the low-level control policies.

\vspace{-2mm}
\subsection{TAMP Environment}
% Tab.~\ref{tab:damp_one_percent_data} and
Tab.~\ref{tab:damp_full_data_all} (appendix) shows planning success rates and VQA performance for the TAMP environment.
The LLM is frozen in these experiments (for pre-trained LLM).
For the results reported in Tab.~\ref{tab:damp_full_data_all}, the input representations are trained on a dataset containing 96,000 training scenes of solely the TAMP environment, \ie no other data is part of the mixture.
For 3-5 objects in the scene, which is the same number as in the training set, most input representations perform similarly well.
However, when increasing the number of objects, it turns out that using a pre-trained LLM improves performance considerably, especially with entity referrals.
Furthermore, we show that a 62B LLM shows better out-of-distribution generalization compared to the 8B variant, while a non-pretrained LLM shows basically no out-of-distribution generalization.
The SayCan baseline \cite{ahn2022can} utilizes oracle affordance functions and has difficulties solving this environment, since affordance functions only constrain what is possible right now, but are not informative enough for the LLM to construct long-horizon plans in TAMP environments.

Tab.~\ref{tab:damp_one_percent_data} shows results for 3-5 objects when training on 1\% of the dataset, which corresponds to only 320 examples for each of the two planning tasks.
Here we see that there are significant differences between the input representations, especially for the planning tasks.
First, pre-training the LLM is beneficial in the low data regime for state inputs.
Second, both ViT variants (ViT+TL, ViT-4B) do not perform well in solving the planning tasks for this little data.
However, if we co-train on all other robot environments as well as general vision-language datasets (ViT-4B generalist), then the performance of the ViT-4B more than doubles.
% , despite only having 320 training examples.
This shows a significant transfer effect between different robot embodiments and tasks.
Finally, using OSRT as the input representation leads to the best performance here, demonstrating the strengths of 3D-aware object representations.
We also observe another instance of transfer here: when we remove the TAMP VQA data and only train on the 640 planning tasks examples, there is a (slight) drop in performance.
The state-of-the art vision-language model PaLI \cite{chen2022pali} that was not trained on robot data is not able to solve the tasks. 
We only evaluated it on $\text{q}_2$ (objects left/right/center on the table) and $\text{q}_3$ (vertical object relations), since those most resemble typical VQA tasks.

\begin{figure}
    \centering
    \includegraphics[width=0.9\columnwidth]{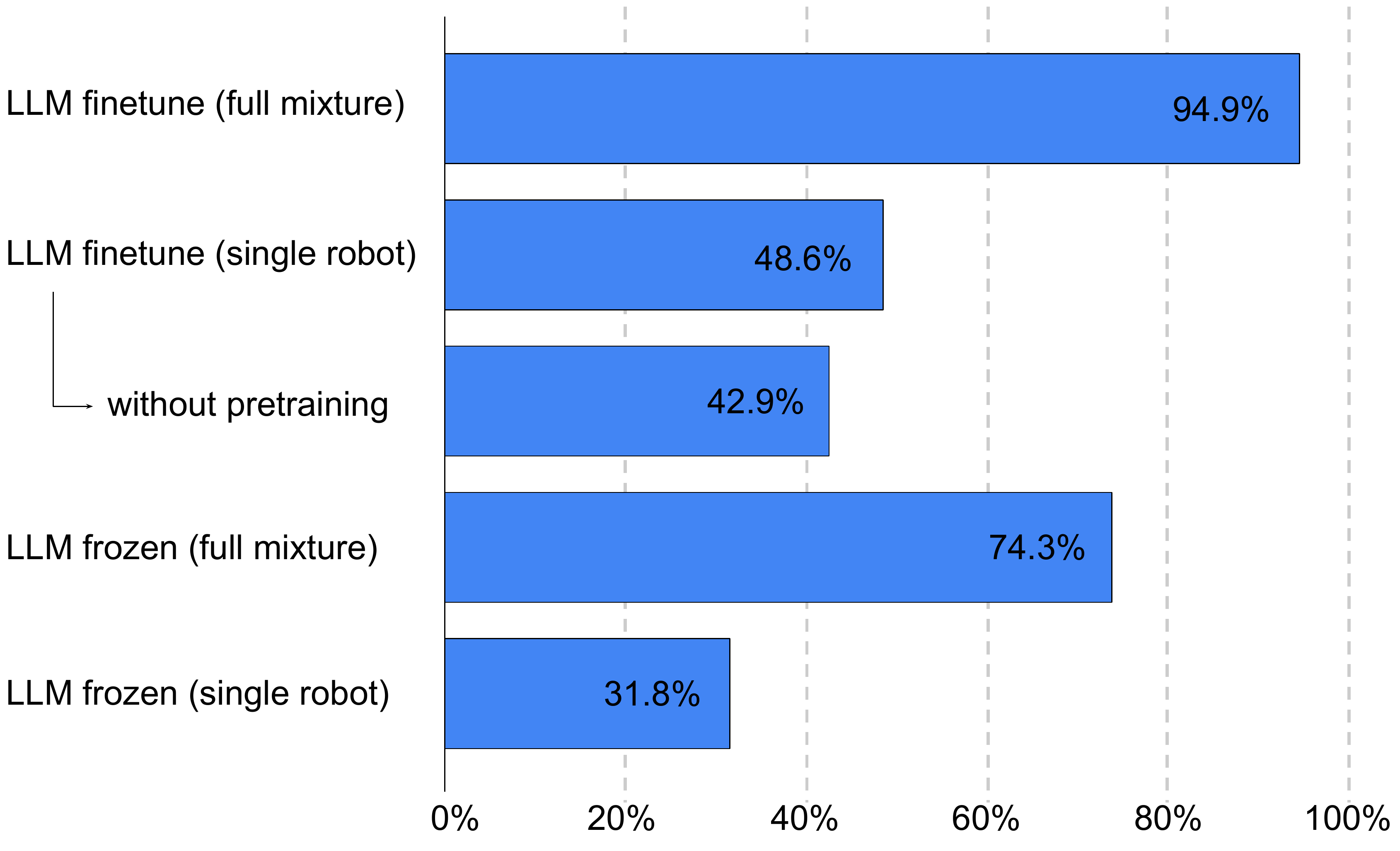}
    \caption{Planning success results in the TAMP environment (1\% data) for PaLM-E-12B, comparing of the effects of PaLM-E models (i) using the full training mixture, (ii) pre-training (ViT and PaLM), and (iii) freezing or finetuning the language model. Transfer from full mixture is particularly effective. Note that full mixture contains only 1\% of the training data (320 examples each) for the tasks evaluated here. Shown is the mean of tasks $\text{p}_1$, $\text{p}_2$.}
    \label{fig:damp_transfer}
\end{figure}

\begin{table}[t]
    \setlength\tabcolsep{2.3pt}
    \resizebox{\columnwidth}{!}{
    \begin{tabular}{lcccccccc}
        \toprule
         & Object- & LLM & \multicolumn{4}{c}{Embodied VQA} & \multicolumn{2}{c}{Planning} \\
         \cmidrule(lr){4-7}\cmidrule(lr){8-9}\\[-5mm]
         & centric & pre-train & $\text{q}_1$ & $\text{q}_2$ & $\text{q}_3$ & $\text{q}_4$ & $\text{p}_1$ & $\text{p}_2$\\
        \midrule
        \multicolumn{2}{l}{SayCan (oracle afford.) \cite{ahn2022can}} & \cmark & - & - & - & - & 38.7 & 33.3 \\
        \multicolumn{2}{l}{PaLI (zero-shot) \cite{chen2022pali}} & \cmark & - & 0.0 & 0.0 & - & - & - \\
        \hline
        \multicolumn{2}{l}{\textit{PaLM-E} (ours) w/ input enc:}\\
        \quad State & \cmark (GT) & \xmark & 99.4 & 89.8 & 90.3 & 88.3 & 45.0 & 46.1  \\
        \quad State & \cmark (GT) & \cmark & \textbf{100.0} & 96.3 & 95.1 & 93.1 & 55.9 & 49.7 \\
        \quad ViT + TL & \cmark (GT) & \cmark & 34.7 & 54.6 & 74.6 & 91.6 & 24.0 & 14.7 \\
        \quad ViT-4B single robot & \xmark & \cmark & - & 45.9 & 78.4 & 92.2 & 30.6 & 32.9 \\
        \quad ViT-4B full mixture & \xmark & \cmark & - & 70.7 & 93.4 & 92.1 & 74.1 & 74.6 \\
        \quad OSRT (no VQA) & \cmark & \cmark & - & - & - & - & 71.9 & 75.1 \\
        \quad OSRT & \cmark & \cmark & 99.7 & \textbf{98.2} & \textbf{100.0} & \textbf{93.7} & \textbf{82.5} & \textbf{76.2} \\
        \bottomrule
    \end{tabular}
    }
    \vspace{-2mm}
    \caption{
    Comparison of different input representations on TAMP environment (in terms of success rates), where data from TAMP constitutes only 1\% (\ie, 320 samples for $\text{p}_1$, $\text{p}_2$ each) of total training data size.
    PaLM-E outperforms both PaLI and SayCan on embodied VQA and planning tasks.
    % respectively
    Cross-domain \textit{transfer} is observed, since the PaLM-E with ViT-4B trained on our full data mixture improves planning performance.  OSRT, despite using no large-scale data, provides the most effective input encodings for learning.
    (GT) means ground-truth object-centric information provided. In all experiments, the LLM is frozen. The non-object centric ViT-4B variant utilizes color to reference objects, hence $\text{q}_1$ cannot be evaluated here. The LLM is frozen in these experiments (except for the case where it is not pre-trained). Sec.~\ref{sec:app:TAMP} describes the tasks $\text{q}_1$-$\text{q}_4$, $\text{p}_1$, $\text{q}_2$.
    }
    \vspace{-4mm}
    \label{tab:damp_one_percent_data}
\end{table}

\begin{table*}[t]
\setlength\tabcolsep{3.4pt}
\parbox{.70\linewidth}{
\centering
\resizebox{\linewidth}{!}{\scriptsize
    \begin{tabular}{lcccccccccccccc}
        \toprule
        \multicolumn{4}{l}{\textit{Zero-shot Baselines}}  &  & & \multicolumn{3}{c}{Task 1} & \multicolumn{3}{c}{Task 2} & \multicolumn{3}{c}{Task 3} \\
        \cmidrule(lr){7-9} \cmidrule(lr){10-12} \cmidrule(lr){13-15}
        \multicolumn{4}{l}{SayCan (oracle afford.) \cite{ahn2022can}} & & & & 0.0 & & &- & & &- \\
        \multicolumn{4}{l}{PaLI \cite{chen2022pali}} &  & & & 0.0 & & &- & & &-\\
        \midrule
                         & trained & from & LLM+ViT & LLM & Task & \multicolumn{3}{l}{\textit{\# Demos}}   \\
        \textit{PaLM-E-} & on & scratch & pretrain & frozen & finetune  & \textit{10} & \textit{20} & \textit{40} & \textit{10} & \textit{20} & \textit{40} & \textit{10} & \textit{20} & \textit{80}\\
        \midrule
        12B & Single robot & \cmark  & \xmark & n/a    & \cmark & 20.0 & 30.0 & 50.0 & 2.5 & 6.3 & 2.5  & 11.3 & 16.9 &   28.3   \\
        12B & Full mixture & \xmark  & \cmark & \cmark & \xmark &  -    &   -   & 20.0 &  -   &  -   & 36.3 &  -   &  -& 29.4 \\
        12B & Full mixture & \xmark  & \cmark & \xmark & \xmark &   -   &   -   & 80.0 & -    & -& 57.5 & -    &   - & 50.0 \\
        12B & Full mixture & \xmark  & \cmark & \xmark & \cmark & \textbf{70.0} & \textbf{80.0} & 80.0 & \textbf{31.3} & \textbf{58.8} & \textbf{58.8} & \textbf{57.5} & \textbf{54.4} & 56.3 \\
        84B & Full mixture & \xmark  & \cmark & \xmark & \xmark &   -   &   -  & \textbf{90.0} & - & - & 53.8 & - & - & \textbf{64.4} \\
        \bottomrule
    \end{tabular}
    }
    \vspace{-3mm}
\caption{Results on planning tasks in the simulated environment from  \citet{lynch2022interactive}.}
\label{tab:lt_sim}
}
\hfill
\parbox{.30\linewidth}{
\centering
{\renewcommand{\arraystretch}{1.08}
\scriptsize
    \vspace{-1mm}
    \begin{center}
    \begin{tabular}{l}
        \toprule
        \textbf{Task 1.} Q: There is a block that is closest to \\
        \textit{\{\ie, top right corner\}}. Push that block to \\
        the other block of the same color. \\
        \midrule
        \textbf{Task 2.} Q: How to sort the blocks by colors \\
        into corners? \\
        \midrule
        \textbf{Task 3.} Q: How to push all the blocks that \\
        are on the \textit{\{left/right\}} side together, \\
        without bringing over any of the blocks \\
        that are on the \textit{\{right/left\}} side? \\
        \bottomrule
    \end{tabular}
    \end{center}
    }
    \vspace{-4mm}
\caption{Task prompts for \cref{tab:lt_sim}.}
}
\vspace{-1em}
\end{table*}

\vspace{-2mm}
\subsection{Language-Table Environment}
\vspace{-1mm}
Tab.~\ref{tab:lt_sim} reports success rates on long-horizon tasks from the Language-Table environment \cite{lynch2022interactive}.
\mbox{PaLM-E} is integrated into a control loop that takes as input the long-horizon task and the current image, and outputs an instruction for the low-level policy.
We see that joint training on internet-scale vision and language results in a more effective model for robot planning, particularly in the few-shot regime with only 10 demos per task.
Scaling the 12B model to the 84B model leads to improvements on 2 of 3 tasks.
As with the TAMP environment, neither SayCan nor zero-shot PaLI are effective, unable to solve the easiest task tested.

\begin{figure*}[h]
  \centering
  \includegraphics[width=0.99\linewidth]{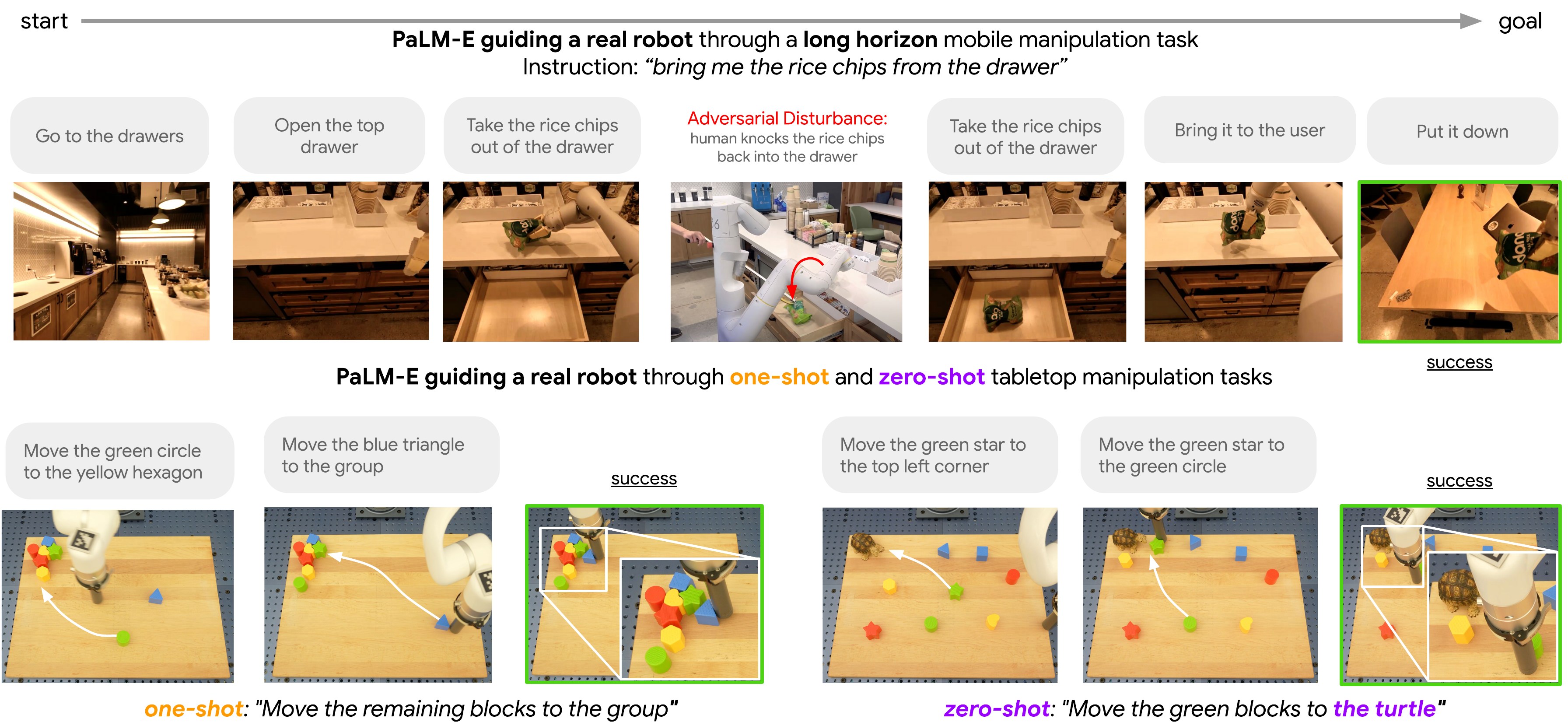}  
 %\vspace{-1.0em}
  \caption{\small{
A single PaLM-E model directs the low-level policies of two real robots. Shown is a long-horizon mobile manipulation task in a kitchen, and one-shot / zero-shot generalization with a tabletop manipulation robot.
  }}
  %\vspace{-1.0em}
  \label{fig:real-fractal}
\end{figure*}

\textbf{Real Robot Results and Few-Shot Generalization.}
In Fig.~\ref{fig:real}, a), we see PaLM-E is capable of guiding a real robot through a multi-stage tabletop manipulation task, while remaining robust to adversarial disturbances. 
Given the observed image and a long-horizon goal, \eg ``sort the blocks by colors into corners", PaLM-E outputs language subgoals at 1 Hz to the policies from \citet{lynch2022interactive}, that output low-level robot actions at 5 Hz. Prior work \cite{lynch2022interactive} instead involved a human in the loop to interactively guide subgoals and corrections.
In \cref{fig:real-fractal}, b) we see PaLM-E is capable of one-shot and zero-shot learning. Here, we finetuned PaLM-E on 100 different long horizon tasks with a single training example each, \eg ``put all the blocks in the center", ``remove the blue blocks from the line". We additionally see that PaLM-E can generalize zero-shot to tasks involving novel object pairs (\cref{fig:real}, c) and to tasks involving objects that were unseen in either the original robot dataset or the finetuning datasets, \eg a toy turtle (\cref{fig:real-fractal}, d).

\vspace{-1mm}
\subsection{Mobile Manipulation Environment}
\vspace{-1mm}
We demonstrate the performance of PaLM-E on challenging and diverse mobile manipulation tasks.
We largely follow the setup in \citet{ahn2022can}, where the robot needs to plan a sequence of navigation and manipulation actions based on an instruction by a human.
For example, given the instruction ``I spilled my drink, can you bring me something to clean it up?", the robot needs to plan a sequence containing ``1. Find a sponge, 2. Pick up the sponge, 3. Bring it to the user, 4. Put down the sponge."
Inspired by these tasks, we develop 3 use cases to test the embodied reasoning abilities of PaLM-E: affordance prediction, failure detection, and long-horizon planning.
The low-level policies are from RT-1~\cite{brohan2022rt}, a transformer model that takes RGB image and natural language instruction, and outputs end-effector control commands.

\textbf{Affordance prediction.}
We investigate PaLM-E's performance at affordance prediction, \ie whether a \texttt{\small skill} of the low-level policy can be executed in the current environment.
This can be formulated as the VQA problem \texttt{\small Given <img>.} \texttt{\small Q: Is it possible to <skill> here?}.
PaLM-E outperforms PaLI (zero-shot), as well as thresholding on value functions trained with QT-OPT (Tab.~\ref{table:fractal_sd}).

\begin{table}
\begin{center}
\resizebox{0.95\columnwidth}{!}{
\begin{tabular}{ l c c c c c c}
\toprule
 \multicolumn{4}{l}{\textit{Baselines}} & Failure det. & Affordance \\ 
 \hline
 \multicolumn{4}{l}{PaLI (Zero-shot) \cite{chen2022pali}} & 0.73 & 0.62 \\
 \multicolumn{4}{l}{CLIP-FT~\cite{xiao2022robotic}} & 0.65 & -  \\
 \multicolumn{4}{l}{CLIP-FT-hindsight~\cite{xiao2022robotic}} & 0.89 & - \\
 \multicolumn{4}{l}{QT-OPT~\cite{kalashnikov2018scalable}} & - & 0.63  \\
 \hline
 \textit{PaLM-E-12B} & from & LLM+ViT & LLM  \\
  trained on & scratch & pretrain & frozen \\
  \midrule
 Single robot & \cmark  & \xmark & n/a    & 0.54 & 0.46 \\
 Single robot & \xmark  & \cmark & \cmark &  \textbf{0.91} & 0.78 \\
 Full mixture & \xmark  & \cmark & \cmark &  \textbf{0.91} & 0.87 \\
 Full mixture & \xmark  & \cmark & \xmark &  0.77 & \textbf{0.91} \\
\bottomrule
\end{tabular}
}
\end{center}
\vspace{-3mm}
\caption{Mobile manipulation environment: failure detection and affordance prediction (F1 score).}
\label{table:fractal_sd}
\vspace{-3mm}
\end{table}

\textbf{Failure detection.}
For a robot to do closed-loop planning, it is also important to detect failures, as is shown in ~\cite{huang2022inner}.
The multi-modal prompt is \texttt{\small Given <img>. Q: Was <skill> successful?}.
Tab.~\ref{table:fractal_sd} shows that PaLM-E outperforms PaLI (zero-shot), as well as a fine-tuned version of CLIP on this dataset.
PaLM-E also outperforms the algorithm proposed in \citet{xiao2022robotic} that leverages two CLIP models trained with hindsight relabeled data.
This method has access to more information than our method, and was specifically designed to just solve failure detection on this dataset.

\textbf{Real robot results: Long-horizon planning.}
Finally, we use PaLM-E to perform \emph{embodied planning} end-to-end for mobile manipulation tasks.
The prompt structure for this task is \texttt{\small Human: <instruction> Robot: <step history>. I see <img>}.
PaLM-E is trained to generate the next step of the plan, conditioned on the history of taken steps and the current image observation of the scene.
After each step is decoded, we map them to a low-level policy as defined in \citet{ahn2022can}.
This process is done in an autoregressive manner, until PaLM-E outputs ``terminate".
We train the model by using the runs from ~\cite{ahn2022can}, which contains 2912 sequences.
We qualitatively evaluated the model in a real kitchen and found the model can carry out long-horizon mobile manipulation tasks, even under adversarial disturbances (Fig.~\ref{fig:real-fractal}).

\vspace{-1mm}
\subsection{Performance on General Visual-Language Tasks}
\vspace{-1mm}
Although it is not the focus of our work, we report in Tab.~\ref{table:general-visual-language} results on general vision-language tasks, including OK-VQA \cite{okvqa}, VQA v2 \cite{balanced_vqa_v2} and COCO captioning \cite{coco}.
A single, generalist PaLM-E-562B model achieves the highest reported number on OK-VQA, including outperforming models finetuned specifically on OK-VQA.
Compared to \cite{tsimpoukelli2021multimodal}, PaLM-E achieves the highest performance on VQA v2 with a frozen LLM to the best of our knowledge.
This establishes that PaLM-E is a competitive visual-language generalist, in addition to being an embodied reasoner on robotic tasks.

\begin{table}
\begin{center}
\resizebox{1.0\columnwidth}{!}{
\setlength\tabcolsep{3.4pt}
\begin{tabular}{ l c c c c}
\toprule
       & \multicolumn{2}{c}{VQAv2} & OK-VQA & COCO  \\ 
 Model & test-dev & test-std & val & Karpathy test  \\ 
 \hline
 \multicolumn{4}{l}{\textit{Generalist (one model)}} \\
 PaLM-E-12B & 76.2 & - & 55.5 & 135.0  \\
 PaLM-E-562B & 80.0 & - & \textbf{66.1} & 138.7  \\
 \hline
 \multicolumn{4}{l}{\textit{Task-specific finetuned models}} \\
 Flamingo \cite{alayrac2022flamingo} & 82.0 & 82.1 & 57.8$\dag$ & 138.1 \\
 PaLI \cite{chen2022pali}    & 84.3 & 84.3 & 64.5 & 149.1       \\
 PaLM-E-12B & 77.7  & 77.9 & 60.1 & 136.0 \\
 PaLM-E-66B & -     & -     & 62.9 & -     \\
 PaLM-E-84B & 80.5  & -     & 63.3 & 138.0   \\
 \hline
 \multicolumn{4}{l}{\textit{Generalist (one model), with frozen LLM}} \\
 \cite{tsimpoukelli2021multimodal} & 48.4 & - & - & -  \\
 PaLM-E-12B frozen & 70.3 & - & 51.5 & 128.0  \\
\bottomrule
\end{tabular}
}
\end{center}
\vspace{-1em}
\caption{Results on general visual-language tasks. For the generalist models, they are the same checkpoint across the different evaluations, while task-specific finetuned models use different-finetuned models for the different tasks.  COCO uses Karpathy splits. $\dag$ is 32-shot on OK-VQA (not finetuned).}
\vspace{-1em}
\label{table:general-visual-language}
\end{table}

\begin{figure}[h]
  \centering
  \includegraphics[width=0.9\linewidth]{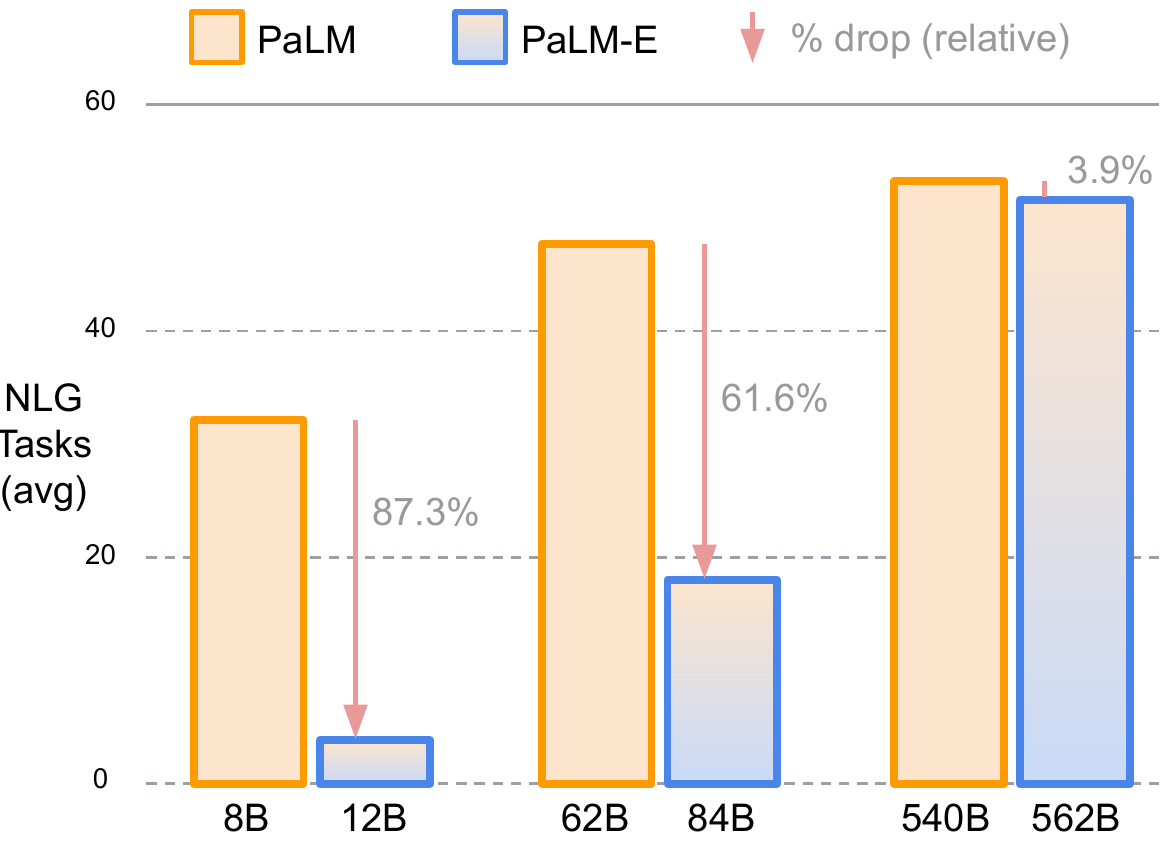}  
 %\vspace{0.0em}
  \caption{\small{
Results on general language tasks (NLG = natural language generation): increasing scale leads to less catastrophic forgetting between a corresponding PaLM-E model and its inherited PaLM model.  See full suite of tasks and results in Tab.~\ref{tab:general-language}.
  }}
  \vspace{-1.0em}
  \label{fig:nlg}
\end{figure}

\vspace{-1mm}
\subsection{Performance on General Language Tasks}
\vspace{-1mm}
Tab.~\ref{tab:general-language} reports the averaged performance of PaLM-E on 21 general language benchmarks for Natural Language Understanding~(NLU) and Natural Language Generation~(NLG) tasks.
The notable trend is that with increasing model scale, there is considerably less catastrophic forgetting of language capabilities. As seen in Fig.~\ref{fig:nlg}, while for the smallest (PaLM-E-12B) model 87.3\% of its NLG performance (relative) has degraded during multimodal training, merely 3.9\% have been degraded for the largest model (PaLM-E-562B).

%%%%%%%%%%%%%%%%%%%%%%%%%%%%%%%%%%%%%%%%%%%%%%%%%%%%%%%%%%%%%%%%%%%%%%%%%%%%%%%%%%%%%%%%%%%%%%%%%%%%%%%%%%%%%%%%%%%%

\vspace{-2mm}
\section{Summary of Experiments \& Discussion}
\vspace{-1mm}

\textbf{Generalist vs specialist models -- transfer.}
As summarized in Fig.~\ref{fig:transfer}, we have shown several instances of \emph{transfer} in this work, meaning that PaLM-E trained on different tasks and datasets at the same time leads to significantly increased performance relative to models trained separately on the different tasks alone.
In Fig.~\ref{fig:damp_transfer}, co-training on the ``full mixture'' achieves more than double the performance.
In Tab.~\ref{table:fractal_sd_app}, we see significant improvements in performance if we add LLM/ViT pre-training, and training on the full mixture instead of the mobile manipulation data alone.
For the Language-Table experiment in Tab.~\ref{tab:lt_sim}, we observe analogous behaviour.

\textbf{Data efficiency.}
Compared to available massive language or vision-language datasets, robotics data is significantly less abundant.
As discussed in the last paragraph, our model exhibits transfer, which
aids PaLM-E to solve robotics tasks from very few training examples in the robotics domain, e.g.\ between 10 and 80 for Language Table or 320 for TAMP.
The OSRT results show another instance of data-efficiency by using a geometric input representation.
A promising opportunity for future work is to combine this with a method benefitting from large-scale visual data.

\textbf{Retaining language capabilities.}
We have shown two paths to retain the language capabilities of the model during multimodal training.
As one option, freezing the LLM and only training the input encoders is a viable path for building embodied language models, although this approach occasionally struggled for robotics tasks (Tab.~\ref{tab:lt_sim}).
As an alternative route, when the whole model is trained end-to-end, the model retains significantly more of its original language performance with increasing model scale (Fig.~\ref{fig:nlg}).

%%%%%%%%%%%%%%%%%%%%%%%%%%%%%%%%%%%%%%%%%%%%%%%%%%%%%%%%%%%%%%%%%%%%%%%%%%%%%%%%%%%%%%%%%%%%%%%%%%%%%%%%%%%%%%%%%%%%

\vspace{-2mm}
\section{Conclusion}
\vspace{-1mm}
We proposed to build an embodied language model by injecting multi-modal information such as images into the embedding space of a pre-trained LLM.
Experiments showed that off-the-shelf state-of-the-art vision-language models trained on general VQA and captioning tasks are not sufficient for embodied reasoning tasks, as well as limitations of a recent proposal for grounding language models through affordances.
To overcome these limitations, we proposed PaLM-E, a single model that is able to control different robots in simulation and in the real world, while at the same time being quantitatively competent at general VQA and captioning tasks. In particular the novel architectural idea of ingesting neural scene representations (\ie, OSRT) into the model is particularly effective, even without large-scale data.
PaLM-E is trained on a mixture of diverse tasks across multiple robot embodiments as well as general vision-language tasks.
Importantly, we have demonstrated that this diverse training leads to several avenues of {\em{transfer}} from the vision-language domains into embodied decision making, enabling robot planning tasks to be achieved data efficiently.
While our results indicate that frozen language models are a viable path towards general-purpose embodied multimodal models that fully retain their language capabilities, we have also surfaced an alternative route with unfrozen models: scaling up the language model size leads to significantly less catastrophic forgetting while becoming an embodied agent.
Our largest model, PaLM-E-562B, showcases emergent capabilities like multimodal chain of thought reasoning, and the ability to reason over multiple images, despite being trained on only single-image prompts.

%\cleardoublepage

\section*{Acknowledgements}
The authors would like to thank, for their advice, help and support: Xi Chen, Etienne Pot, Sebastian Goodman, Maria Attarian, Ted Xiao, Keerthana Gopalakrishnan, Kehang Han, Henryk Michalewski,  Neil Houlsby, Basil Mustafa, Justin Gilmer, Yonghui Wu, Erica Moreira, Victor Gomes, Tom Duerig, Henning Meyer, and Kendra Byrne.

\bibliography{example_paper}
\bibliographystyle{icml2022}

%%%%%%%%%%%%%%%%%%%%%%%%%%%%%%%%%%%%%%%%%%%%%%%%%%%%%%%%%%%%%%%%%%%%%%%%%%%%%%%
%%%%%%%%%%%%%%%%%%%%%%%%%%%%%%%%%%%%%%%%%%%%%%%%%%%%%%%%%%%%%%%%%%%%%%%%%%%%%%%
% APPENDIX
%%%%%%%%%%%%%%%%%%%%%%%%%%%%%%%%%%%%%%%%%%%%%%%%%%%%%%%%%%%%%%%%%%%%%%%%%%%%%%%
%%%%%%%%%%%%%%%%%%%%%%%%%%%%%%%%%%%%%%%%%%%%%%%%%%%%%%%%%%%%%%%%%%%%%%%%%%%%%%%
\newpage
\appendix
\onecolumn

\begin{figure*}[h!]
  \centering
  \includegraphics[width=0.99\linewidth]{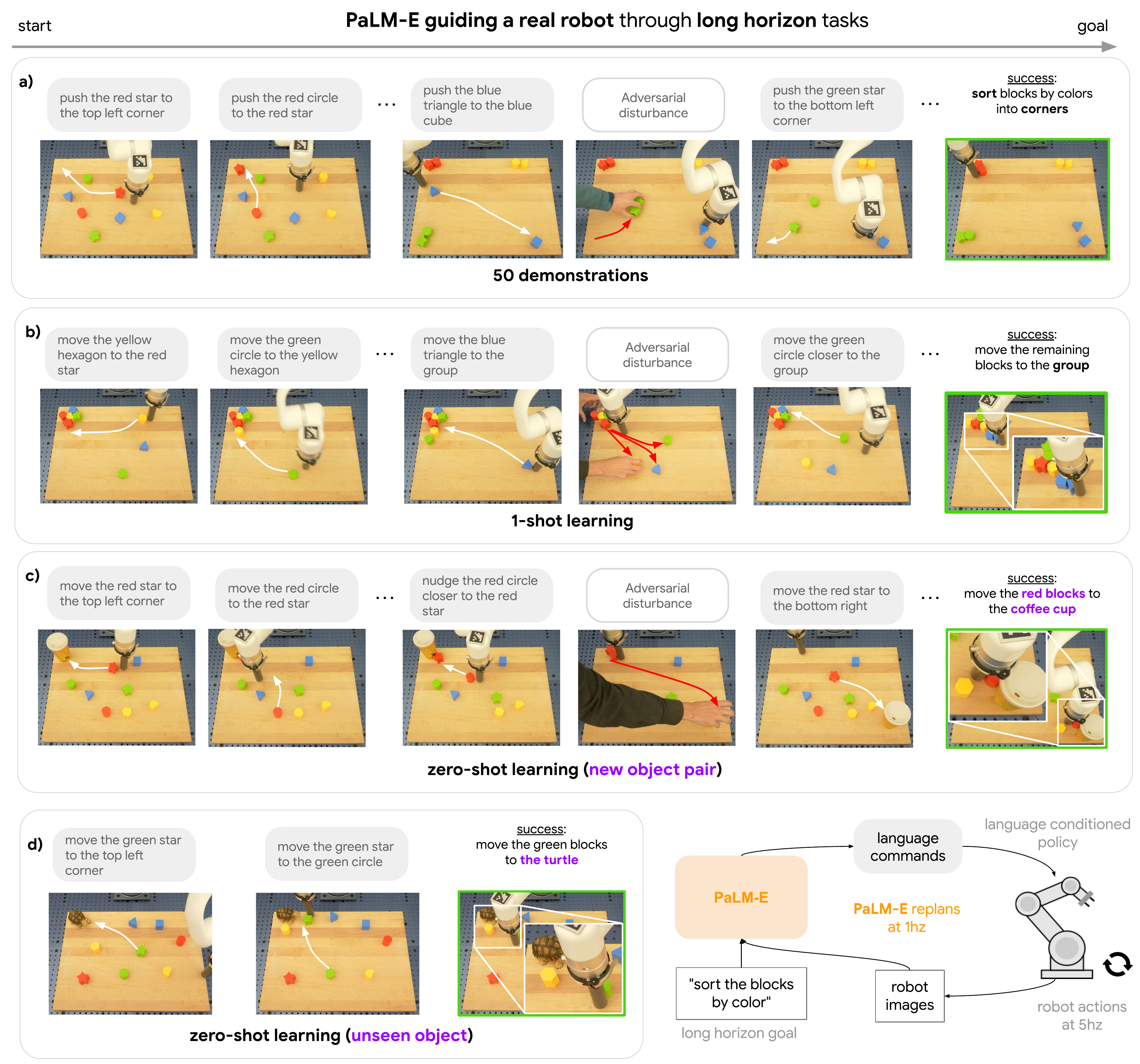}  
  \caption{\small{
  PaLM-E interactively guides a real robot through long-horizon manipulation tasks on Language-Table, while remaining robust to adversarial disturbances. We find evidence that PaLM-E is capable of one-shot and zero shot generalization.
  }}
  \vspace{-1.0em}
  \label{fig:real}
\end{figure*}

\section{Data Mixture}\label{sec:app:dataMixture}
Tab.~\ref{tab:full-mixture} shows the dataset and sampling frequency for the ``full mixture'' as referred to in the experiments.
The majority of the data distribution is general vision-language tasks, with less than 10\% robot data.

\begin{table}[!htp]\centering
\begin{tabular}{lrrr}\toprule
Dataset in full mixture &Sampling frequency &\% \\\midrule
Webli \cite{chen2022pali} &100 &52.4 \\
VQ$^2$A \cite{vq2a} &25 &13.1 \\
VQG \cite{vq2a} &10 &5.2 \\
CC3M \cite{sharma2018conceptual} &25 &13.1 \\
Object Aware \cite{object_aware} &10 &5.2 \\
OKVQA \cite{okvqa} &1 &0.5 \\
VQAv2 \cite{balanced_vqa_v2} &1 &0.5 \\
COCO \cite{coco} &1 &0.5 \\
Wikipedia text &1 &0.5 \\
(robot) Mobile Manipulator, real &6 &3.1 \\
(robot) Language Table \cite{lynch2022interactive}, sim and real &8 &4.2 \\
(robot) TAMP, sim &3 &1.6 \\
\bottomrule
\end{tabular}
\caption{Dataset sampling frequency and ratio for the ``full mixture'' referred to in experiments.}\label{tab:full-mixture}
\end{table}

\section{Environment Details}
\subsection{Task and Motion Planning (TAMP)}\label{sec:app:TAMP}
\begin{figure}
    \centering
    \includegraphics[width=4cm]{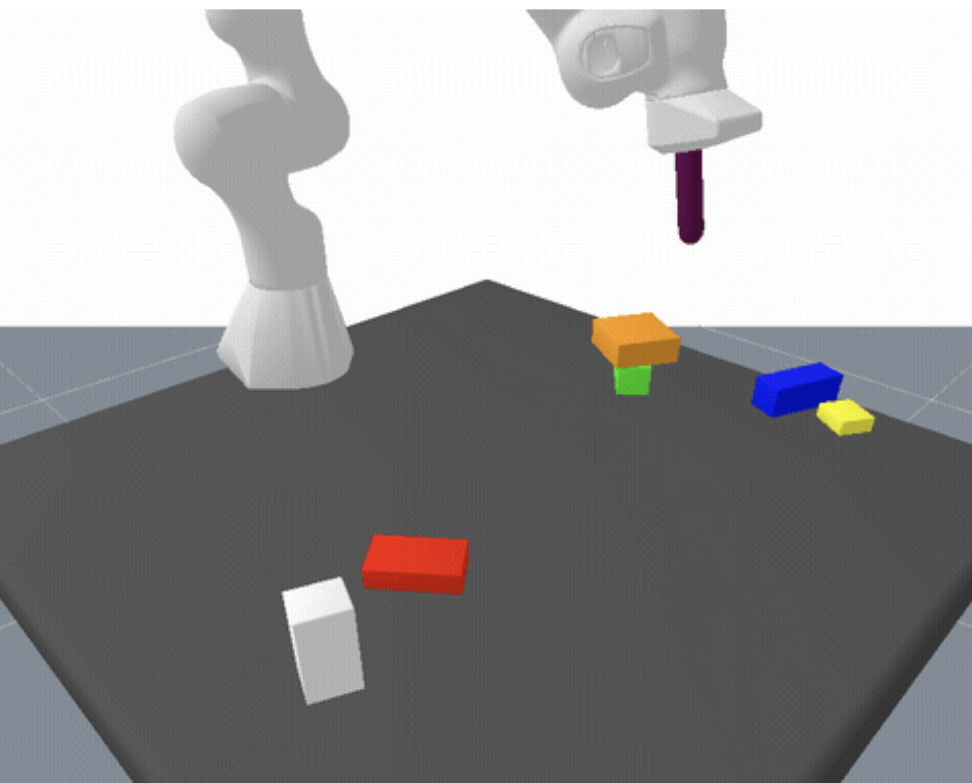}
    \includegraphics[width=4cm]{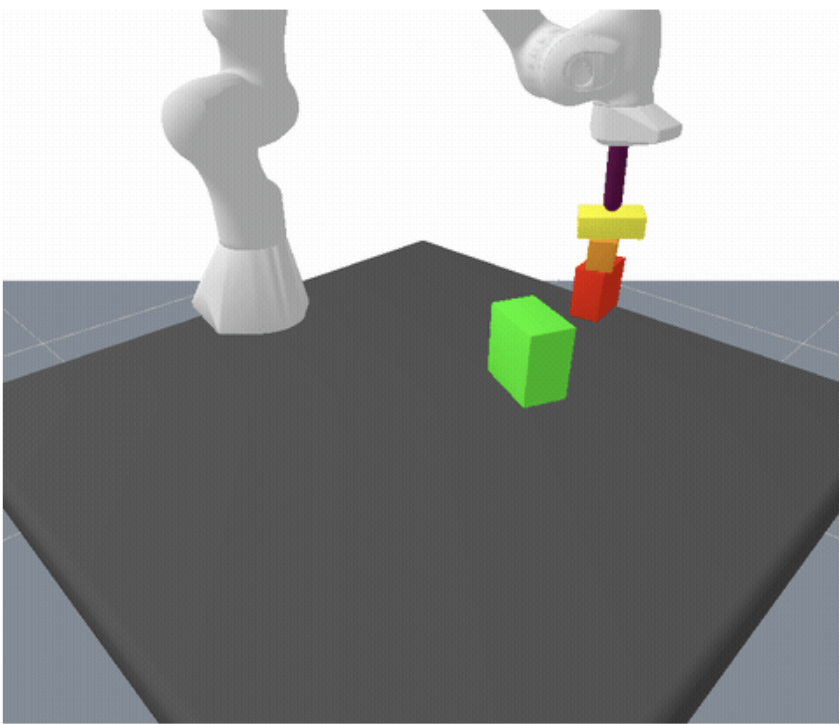}
    \caption{Two TAMP environment test examples. Left with 6 objects (training data contains 3-5 objects), right with 4 objects.}
    \label{fig:app:damp}
\end{figure}

The training scenes for the TAMP environment contain 3-5 cube-shaped objects of different sizes, colors and sampled initial poses.
Fig.~\ref{fig:app:damp} show an example test scene that contains 6 objects.

In the global version, we consider the following three VQA tasks:
\begin{itemize}
    \item $\text{q}_2$: object-table relation. Example prompt: \texttt{Given <img>. Q: Is the red object left, right, or center of the table?}. Target: \texttt{A: The red object is in the center of the table.}
    \item $\text{q}_3$: object-object relations. Example prompt: \texttt{Given <img>. Q: Is the yellow object below the blue object?}. Target: \texttt{A: No, the yellow object is not below the blue object.}
    \item $\text{q}_4$: plan feasibility. Example prompt: \texttt{Given <img>. Q: Is it possible to first grasp the blue object, then place it on the yellow object, and then grasp the yellow object?}. Target: \texttt{A: No, this is not possible.}
\end{itemize}
as well as the two planning tasks
\begin{itemize}
    \item $\text{p}_1$: grasping. Example prompt: \texttt{Given <img>. Q: How to grasp the green object?}. Target: \texttt{A: First grasp the orange object and place it on the table, then grasp the green object.}
    \item $\text{p}_2$: stacking. Example prompt: \texttt{Given <img>. Q: How to stack the white object on top of the red object?}. Target: \texttt{A: First grasp the green object and place it on the table, then grasp the white object and place it on the red object.}
\end{itemize}

For the object-centric version with entity referrals, all prompts contain the prefix \texttt{<prefix>} = \texttt{Obj 1 is <obj$_1$>.} $\ldots$ \texttt{Obj j is <obj$_j$>.}, and the VQA task $\text{q}_1$ is about the color of an object. The other tasks (except with the different prefix, and entity referrals), remain the same.

We utilize the planner from \citet{20-driess-RSS} to generate the dataset for the planning tasks.
The low-level policies are also obtained with the method of \citet{20-driess-RSS}.

\subsection{Interactive Language Table}

We use the Language-Table real-world tabletop setup and simulated environment from Interactive Language \cite{lynch2022interactive}.

\textbf{Data collection.} For each task, given the long horizon instruction, we prompt a labeler to enter a short horizon command every 4 seconds. We pass the short horizon instructions to an Interactive Language policy trained using the same procedure as in \citet{lynch2022interactive}. The policy executes 40 steps (10Hz for 4 seconds) before requiring another command from the labeler. This is repeated until the labeler determines the long horizon instruction is complete and issues a 'done' instruction.  The data collection procedure for the real world experiments are the same as in simulation.

\textbf{Train and Evaluation.} To train the finetuned versions of these models, we train a pretrained PaLM-E model for 9,000 additional steps, in order to support a data complexity sweep without training several separate models from scratch on slightly different versions of the full mixture. For Tasks 2 and 3 in simulation, we implement an automated reward to measure the success rate, and we evaluate PaLM-E by running 80 rollouts for each task. Given the current image and high level task, PaLM-E issues a text instruction which a trained low-level policy executes for 4 seconds before PaLM-E issues a new text instruction. For Task 1, we use a test-set and report validation accuracy. This is because the task only requires one step to solve, despite being a complicated visual and linguistic processing task and cannot be solved by the low-level policy from the prompt alone.

\begin{table*}[t] \small
    \centering
    \begin{tabular}{lcccccccc}
        \toprule
         & $\phi$ & LLM pre-trained & $\text{q}_1$ & $\text{q}_2$ & $\text{q}_3$ & $\text{q}_4$ & $\text{p}_1$ & $\text{p}_2$ \\
        \midrule
        \multirow{10}{1cm}{3 - 5 objects} & SayCan (w/ oracle affordances) & \cmark & - & - & - & - & 38.7 & 33.3 \\
        & state & \xmark & 100.0 & 99.3 & 98.5 & 99.8 & 97.2 & 95.5 \\
        & state & \cmark (unfrozen) & 100.0 & 98.8 & 100.0 & 97.6 & 97.7 & 95.3  \\
        & state & \cmark & 100.0 & 98.4 & 99.7 & 98.5 & 97.6 & 96.0  \\
        & state (w/o entity referrals) & \cmark & 100.0 & 98.8 & 97.5 & 98.1 & 94.6 & 90.3 \\
        & ViT + TL (obj.\ centric) & \cmark & 99.6 & 98.7 & 98.4 & 96.8 & 9.2 & 94.5 \\
        & ViT + TL (global) & \cmark & - & 60.7 & 90.8 & 94.3 & 70.7 & 69.2 \\
        %& ViT-4B (obj.\ centric) & \cmark \\
        & ViT-4B (global) & \cmark & - & 98.2 & 99.4 & 99.0 & 96.0 & 93.4 \\
        & ViT-4B generalist & \cmark & - & 97.1 & 100.0 & 98.9 & 97.5 & 95.2 \\
        & OSRT & \cmark & 99.6 & 99.1 & 100.0 & 98.8 & 98.1 & 95.7 \\
        \midrule
        \multirow{3}{1cm}{6 objects} & state & \xmark & 20.4 & 39.2 & 71.4 & 85.2 & 56.5 & 34.3 \\
        & state & \cmark & 100.0 & 98.5 & 94.0 & 89.3 & 95.3 & 81.4 \\
        & state (w/o entity referrals) & \cmark & 77.7 & 83.7 & 93.6 & 91.0 & 81.2 & 57.1 \\
        \midrule
        \multirow{3}{1cm}{8 objects} & state & \xmark & 18.4 & 27.1 & 38.1 & 87.5 & 24.6 & 6.7 \\
        & state & \cmark & 100.0 & 98.3 & 95.3 & 89.8 & 91.3 & 89.3 \\
        & state (w/o entity referrals) & \cmark & 60.0 & 67.1 & 94.1 & 81.2 & 49.3 & 49.3 \\
        \midrule
        \multirow{3}{1.6cm}{6 objects + OOD tasks} & state (8B LLM) & \xmark & - & 0 & 0 & 72.0 & 0 & 0 \\
        & state (8B LLM) & \cmark & - & 49.3 & 89.8 & 68.5 & 28.2 & 15.7 \\
        & state (62B LLM) & \cmark & - & 48.7 & 92.5 & 88.1 & 40.0 & 30.0 \\
        \bottomrule
    \end{tabular}
    \caption{Success rates on TAMP environment for different input representations. 3-5 objects in the scene correspond to the training distribution. OOD tasks means out-of-distribution tasks where the objects are referenced by color, although in the trainig data they have been referenced by their special tokens \texttt{\small obj$_j$} in the object-centric case. The SayCan baseline \cite{ahn2022can} utilizes oracle, one-step affordance functions.}
    \label{tab:damp_full_data_all}
\end{table*}

%\begin{table*}[t]
%    \centering
%    \begin{tabular}{ccccccccc}
%        \toprule
%        $\phi$ & obj.-centric & LLM pre-trained & c1 & c2 & c3 & c4 & p1 & p2\\
%        \midrule
%        \multicolumn{2}{c}{SayCan (w/ oracle affordances)} & \cmark & & & & & 38.7 & 33.3 \\
%        state & \cmark (GT) & \xmark & 99.4 & 89.8 & 90.3 & 88.3 & 45.0 & 46.1  \\
%        state & \cmark (GT) & \cmark & 100.0 & 96.3 & 95.1 & 93.1 & 55.9 & 49.7 \\
%        ViT + TL & \cmark (GT) & \cmark & 34.7 & 54.6 & 74.6 & 91.6 & 24.0 & 14.7 \\
%        ViT-4B & \xmark & \cmark & - & 45.9 & 78.4 & 92.2 & 30.6 & 32.9 \\
%        ViT-4B generalist & \xmark & \cmark & - & 70.7 & 93.4 & 92.1 & 74.1 & 74.6 \\
%        OSRT (no VQA) & \cmark & \cmark & - & - & - & - & 71.9 & 75.1 \\
%        OSRT & \cmark & \cmark & 99.7 & 98.2 & 100.0 & 93.7 & 82.5 & 76.2 \\
%        \bottomrule
%    \end{tabular}
%    \caption{Success rates on TAMP environment for only 1\% of original training data size. (GT) means ground-truth object-centric information provided (either in terms of object masks, or factored state). OSRT does not use ground-truth object centric information. In all experiments, the LLM is frozen during training of the encoders $\phi$. The non-object centric ViT-4B variant utilizes color to reference objects, hence c1 cannot be evaluated here. 1\% of the original data means only 320 training samples for p1 and p2 each.}
%    \label{tab:damp_one_percent_data_all}
%\end{table*}

\clearpage

\section{Natural Language Generation and Understanding Results}\label{sec:app:nlgnlu}

\begin{table}[!htp]\centering
\scriptsize
\begin{tabular}{lrr@{\hskip 0.9cm}rr@{\hskip 0.9cm}rr@{\hskip 0.9cm}r}\toprule
\textbf{} &\textbf{PaLM-8B} &\textbf{PaLM-E-12B} &\textbf{PaLM-62B} &\textbf{PaLM-E-84B} & \textbf{PaLM-540B} & \textbf{PaLM-E-562B} & \textbf{Category} \\
\textbf{1-shot evals} & &\textit{(unfrozen)} &  & \textit{(unfrozen)} & & \textit{(unfrozen)} \\\midrule
TriviaQA (wiki) (EM) & 48.5 & 10.1 & 72.7 & 31.8 & 81.4 & 74.6 & NLG \\
Natural Questions (EM) &10.6 &1.6 &23.1 &7.6 &29.3 &27.2 &NLG \\
WebQuestions (EM) &12.6 &3.4 &19.8 &7.9 &22.6 &21.8 &NLG \\
Lambada &57.8 &1.4 &75.5 &26.1 &81.8 &83.3 &NLG \\
HellaSwag &68.2 &48.4 &79.7 &75.3 &83.6 &83.5 &NLU \\
StoryCloze &78.7 &68.7 &83.8 &83.9 &86.1 &86.3 &NLU \\
Winograd &82.4 &71.8 &85.3 &86.4 &87.5 &89.0 &NLU \\
Winogrande &68.3 &55.3 &76.8 &72.5 &83.7 &83.0 &NLU \\
RACE-M &57.7 &43.2 &64.1 &57.4 &69.3 &70.3 &NLU \\
RACE-H &41.6 &33.2 &48.7 &42.3 &52.1 &52.8 &NLU \\
PIQA &76.1 &68.1 &80.9 &78.2 &83.9 &84.9 &NLU \\
ARC-e &71.3 &53.4 &78.9 &71.4 &85.0 &86.3 &NLU \\
ARC-c &42.3 &30.9 &51.8 &46.7 &60.1 &62.6 &NLU \\
OpenBookQA &47.4 &41.4 &51.2 &51.6 &53.6 &55.8 &NLU \\
BoolQ &64.7 &61.6 &83.1 &81.6 &88.7 &89.4 &NLU \\
Copa &82.0 &77.0 &93.0 &91.0 &91.0 &93.0 &NLU \\
RTE &57.8 &54.9 &71.5 &59.6 &78.7 &75.1 &NLU \\
Wic &50.6 &50.0 &48.6 &50.2 &63.2 &64.1 &NLU \\
WSC &81.4 &68.4 &84.9 &75.8 &86.3 &85.6 &NLU \\
ReCoRD &87.8 &71.2 &91.0 &78.5 &92.8 &92.5 &NLU \\
CB &41.1 &37.5 &55.4 &73.2 &83.9 &80.3 &NLU \\
& & & & & & & \\
Avg NLU &64.7 &55.0 &72.3 &69.2 &78.2 &78.5 & \\
Avg NLG &32.4 &4.1 &47.8 &18.4 &53.8 &51.7 & \\
& & & & & & & \\
NLU delta (\%, relative) & &-15.0\% & &-4.3\% & &+0.4\% & \\
NLG delta (\%, relative) & &-87.3\% & &-61.6\% & &-3.8\% & \\
\bottomrule
\end{tabular}
\caption{Full language evaluation task results on both NLU and NLG tasks, for both the original PaLM models and for associated PaLM-E (unfrozen) models.
The PaLM-E models with a frozen LLM have the same performance as their corresponding underlying PaLM models.
}\label{tab:general-language}
\end{table}

\clearpage

\section{Additional Data for Affordance and Success Detection}

\begin{table}[h]
\begin{center}
\scriptsize
\begin{tabular}{ l c c c c c c}
\toprule
 \multicolumn{4}{l}{Model} & Precision & Recall & F1-score \\ 
 \hline
 \multicolumn{4}{l}{PaLI (Zero-shot) \cite{chen2022pali}} & 0.59  & 0.98  & 0.73 \\
 \multicolumn{4}{l}{CLIP-FT~\cite{xiao2022robotic}} & 0.50 & 0.95 & 0.65  \\
 \multicolumn{4}{l}{CLIP-FT-hindsight~\cite{xiao2022robotic}} & 1.0 & 0.80 & 0.89  \\
 \hline
 \textit{PaLM-E-12B} & from & LLM+ViT & LLM  \\
  trained on & scratch & pretrain & frozen \\
  \midrule
 Single robot & \cmark  & \xmark & n/a    &  0.52 & 0.55 & 0.54 \\
 Single robot & \xmark  & \cmark & \cmark &  0.91     & 0.92     & \textbf{0.91} \\
 Full mixture & \xmark  & \cmark & \cmark &  0.89     & 0.93 & \textbf{0.91} \\
 Full mixture & \xmark  & \cmark & \xmark &  0.66 & 0.91 & 0.77 \\
 % PaLM-E-12B & 0.91 & 0.87 &  \textbf{0.89} \\
 %PaLM-E-84B & x & y & z \\
\bottomrule
\end{tabular}
\end{center}
\vspace{-1em}
\caption{Mobile manipulation environment: failure detection, showing individual precision and recall scores.}
\label{table:fractal_sd_app}
\end{table}

\begin{table}[h]
\begin{center}
\scriptsize
\begin{tabular}{ l c c c c c c}
\toprule
 \multicolumn{4}{l}{Model} & Precision & Recall & F1-score \\ 
 \hline
 \multicolumn{4}{l}{PaLI (Zero-shot) \cite{chen2022pali}} & 0.57  & 0.69  & 0.62 \\
 \multicolumn{4}{l}{QT-OPT~\cite{kalashnikov2018scalable}} & 0.60 & 0.67 & 0.63  \\
 \hline
 \textit{PaLM-E-12B} & from & LLM+ViT & LLM  \\
  trained on & scratch & pretrain & frozen \\
  \midrule
 Single robot & \cmark  & \xmark & n/a    &  0.67 & 0.35 & 0.46 \\
 Single robot & \xmark  & \cmark & \cmark &  0.90 & 0.69     & 0.78 \\
 Full mixture & \xmark  & \cmark & \cmark &  0.95     & 0.80 & 0.87 \\
 Full mixture & \xmark  & \cmark & \xmark &  0.92 & 0.88 & \textbf{0.91} \\
\bottomrule
\end{tabular}
\end{center}
\vspace{-1em}
\caption{Mobile manipulation environment: affordance prediction, showing individual precision and recall scores.}
\label{table:fractal_affordance_app}
\end{table}

\section{Image Attribution}\label{sec:app:imgAttribution}

The image of the New York Knicks and Boston Celtics in Figure 2 is under the terms CC-by-2.0 (\url{https://creativecommons.org/licenses/by/2.0/}), and was posted to Flickr by kowarski at \url{https://www.flickr.com/photos/27728232@N00/8666371367}.  The egocentric video images are from \url{https://youtu.be/-UXKmqBPk1w}, as in \cite{zeng2022socratic}, via permission from creator Cody Wanner.

%%%%%%%%%%%%%%%%%%%%%%%%%%%%%%%%%%%%%%%%%%%%%%%%%%%%%%%%%%%%%%%%%%%%%%%%%%%%%%%
%%%%%%%%%%%%%%%%%%%%%%%%%%%%%%%%%%%%%%%%%%%%%%%%%%%%%%%%%%%%%%%%%%%%%%%%%%%%%%%

\end{document}